\documentclass[runningheads]{llncs}
\usepackage{graphicx}
\usepackage{booktabs}
\usepackage{multirow}
\usepackage{eurosym}
\usepackage{hyperref}
\usepackage[noend]{algpseudocode}
\usepackage{algorithm}
\usepackage{arydshln}
\usepackage{amsmath}
\usepackage[para]{footmisc}
\usepackage{subcaption}
\usepackage{xspace}

\usepackage{orcidlink}

\usepackage{todonotes}

\usepackage{cleveref}

\usepackage{pdflscape}

\usepackage[inline]{enumitem}

\newcommand*{\tran}{^{\mkern-1.5mu\mathsf{T}}}

\newcommand{\fscoreII}{F-score\textsubscript{2}\xspace}

\newcommand{\etal}{\textit{et al.}}

\begin{document}
\title{Linear Object Detection in Document Images using Multiple Object Tracking}
\titlerunning{Linear Object Detection using MOT}
\author{%
Philippe Bernet%
\and
Joseph Chazalon\orcidlink{0000-0002-3757-074X}\and
Edwin Carlinet\orcidlink{0000-0001-5737-5266}\and\\
Alexandre Bourquelot%
\and
Elodie Puybareau\orcidlink{0000-0002-2748-6624}%
}
\authorrunning{%
P. Bernet et al.
}
\institute{%
EPITA Research Lab (LRE), Le Kremlin-Bicêtre, France\\
\email{\{firstname.lastname\}@epita.fr}%
}
\maketitle              %
\begin{abstract}

Linear objects convey substantial information about document structure, but are challenging to detect accurately because of degradation (curved, erased) or decoration (doubled, dashed).
Many approaches can recover some vector representation, but only one closed-source technique introduced in 1994, based on Kalman filters (a particular case of Multiple Object Tracking algorithm), can perform a pixel-accurate instance segmentation of linear objects and enable to selectively remove them from the original image.
We aim at re-popularizing this approach and propose: 1. a framework for accurate instance segmentation of linear objects in document images using Multiple Object Tracking (MOT); 2. document image datasets and metrics which enable both vector- and pixel-based evaluation of linear object detection; 3. performance measures of MOT approaches against modern segment detectors; 4. performance measures of various tracking strategies, exhibiting alternatives to the original Kalman filters approach; and 5. an open-source implementation of a detector which can discriminate instances of curved, erased, dashed, intersecting and/or overlapping linear objects.

\end{abstract}

\begin{figure}[!b]
    \centering
    \includegraphics[width=\linewidth]{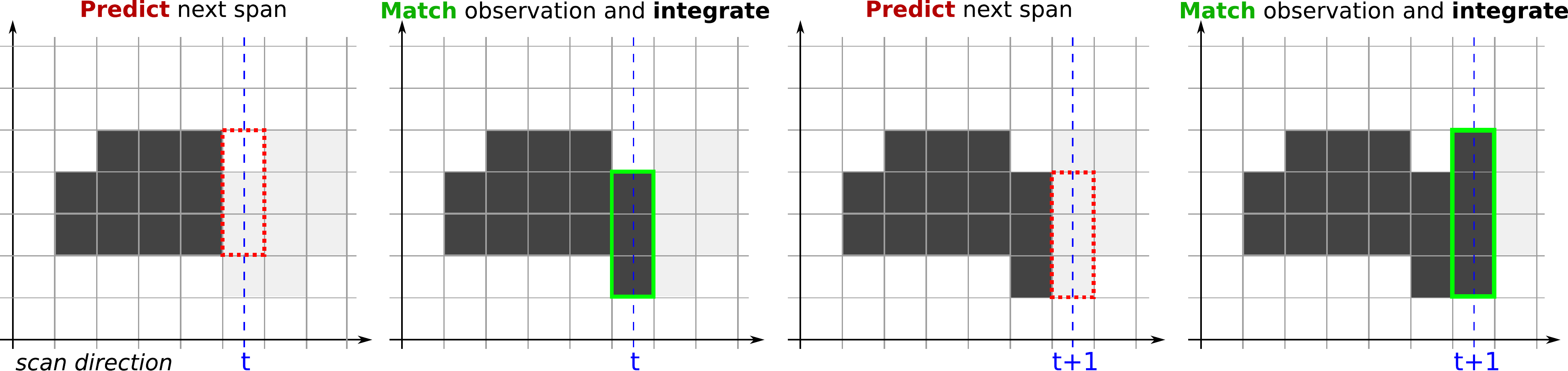}
    \caption{Tracking process of linear objects.
    Images are scanned in a single left-to-right (resp. top-to-bottom) pass.
    At each step $t$, the tracker predicts the next \emph{span} (red box) of a linear object from past (opaque) observations,
    then it matches candidate observations, selects the most probable one (green box),
    and adjust its internal state accordingly.
    Multiple Object Tracking (MOT) handles intersections, overlaps, gaps, and detects boundaries accurately.
    }
    \label{fig:mot_line_detection}
\end{figure}

\section{Introduction}
\label{sec:introduction}
The detection of linear structures is often cast as a boundary detection problem in Computer Vision (CV), 
focusing on local gradient maxima separating homogeneous regions to reveal edges in natural images.
Document images, however, often exhibit very contrasted and thin strokes which carry main information, containing printings and writings, overlaid on some homogeneous ``background''.
It is therefore common to observe a local maximum and a local minimum in a short range along the direction normal to such stroke.
As a result, methods based on region contrast tend to fail on document images, rejecting text and strokes as noise or produce double-detections around linear structures.%

\begin{figure}[tb]
    \centering
    \setlength{\tabcolsep}{5pt}
    \begin{tabular}{ccc}
          \fbox{\includegraphics[width=30mm]{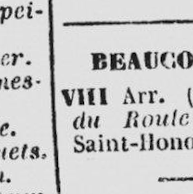}} &
          \fbox{\includegraphics[width=30mm]{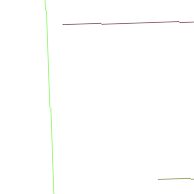}} &
          \fbox{\includegraphics[width=30mm]{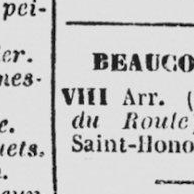}}\\[2mm]
         \fbox{\includegraphics[width=30mm]{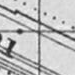}} &
         \fbox{\includegraphics[width=30mm]{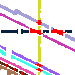}} &
         \fbox{\includegraphics[width=30mm]{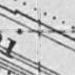}}\\
    \end{tabular}
    
    \caption{Two possible applications. %
    \emph{First row} illustrates an orientation detection task,
    relying on locating endpoints of vertical lines.
    \emph{Second row} illustrates a multi-stage information extraction task,
    relying on discriminating between building shapes and georeferencing lines (horizontal and vertical) in map images.
    \emph{Left}: image inputs, \emph{center}: detections, \emph{right}: rotated or inpainted result.
    } 
    \label{fig:taskillustration}
\end{figure}

In this work, we are interested in enabling a fast and accurate pre-processing which can detect and eventually remove decorated, degraded or overlapping linear objects in document images.
\Cref{fig:taskillustration} illustrates the two major outputs such method needs to produce: a vector output containing the simplified representation of all start and end coordinates of approximated line segments; and a pixel-accurate instance segmentation where each pixel is assigned zero (background), one, or more (intersections) linear object label(s), enabling their removal while preserving the integrity of other shapes (i.e. without creating gaps).

Our work aims at reviving a technique proposed in 1994~\cite{dandecy_kalman_1994,leplumey_kalman_1995,poulain_dandecy_analyse_1995} which introduced a way to leverage Kalman filters to detect and segment instances of linear objects in document images, for which no public implementation is available.
We propose a re-construction of this method with the following new contributions:
\begin{enumerate*}
    \item we introduce a general framework for accurate instance segmentation of linear objects in document images using Multiple Object Tracking (MOT), a more general approach, which enables to separate the stages of the approach properly and eventually leverage deep image filters, and also to extend the method with new trackers;
    \item we introduce datasets and metrics which enable both vector- and pixel-based evaluation of linear object detector on document images: from existing and newly-created contents we propose vector and pixel-accurate ground-truth for linear object and release it publicly;
    \item we show the performance of MOT approaches against modern segment detectors on a reproducible benchmark using our public datasets and evaluation code;
    \item we compare the performance of various tracking strategies, exhibiting alternatives to the original Kalman filters approach and introducing, in particular, a better parametrization of the Kalman filters;
    and 
    \item we release the first open-source implementation %
    of a fast, accurate and extendable detector which can discriminate instances of curved, dashed, intersecting and/or overlapping linear objects.%
\end{enumerate*}

This study is organized as follows.
We review existing relevant methods for linear object detection in \Cref{sec:sota},
detail the formalization of our proposed framework in \Cref{sec:framework},
and present the evaluation protocol (including the presentation of datasets and metrics), as well as the results of our benchmarks for vector and pixel-accurate detection, in \Cref{sec:experiments}.

\section{State of the Art in Linear Object Detection}
\label{sec:sota}

The two complementary goals addressed in this work were traditionally tackled by distinct method categories.
Among these categories, some produce vector outputs
(with end coordinates, but also sometimes sparse or dense coordinates of points laying on the object),
and some others provide some pixel labeling of linear objects.
Such pixel labeling is often limited to a \emph{semantic segmentation}~\cite{kirillov_panoptic_2019}, i.e. a binary ``foreground'' vs ``background'' pixel-wise classification,
and, as we will see, only tracking-based approaches can propose an \emph{instance segmentation}~\cite{kirillov_panoptic_2019}, i.e. effectively assign each pixel to a set of objects.
Due to the lack of datasets for linear object detection in document images,
we are particularly interested, in this current work, in techniques which have light requirements on training data,
and leave for future research the implementation of a larger benchmark. %
Linear object detection is very related to the problem of \emph{wireframe parsing}, well studied in CV, 
for which several datasets were proposed~\cite{denis_efficient_2008,huang_learning_2018},
but their applicability to document images, which exhibit distinct structural and textural properties, were not specifically studied.
We review a selection of approaches which pioneered key ideas in each category of the taxonomy we propose here.

\paragraph{Pixel-wise edge classifiers.}
The Sobel operator~\cite{sobel_isotropic_2015} is a good example of an early approach which predicts,
considering a limited spatial context,
the probability or score that a particular image location belongs to an edge.
This large family includes all hourglass-shaped deep \emph{semantic segmentation} networks,
notably any approach designed around some U-Net variant~\cite{ronneberger_u-net:_2015,lin_feature_2017}
like the HED~\cite{xie_holistically-nested_2015} and BDCN~\cite{he_bi-directional_2019} edge detectors.
Such approaches can make pixel-accurate predictions but cannot discriminate object instances.
To enable their use for vectorization,
Xue et al.~\cite{xue_learning_2019} proposed to predict an Attraction Field Map
which eases the extraction of 1-pixel-thin edges after a post-processing stage.
EDTER~\cite{pu_edter_2022} pushes the \emph{semantic segmentation} approach as far as possible with a transformer encoder, enabling to capture a larger context.
While all these techniques can produce visually convincing results,
they cannot assign a pixel to more than one object label and require a post-processing to produce some vector output.
Furthermore, their training requirements and computational costs forces us to remove them from our comparison,
but they could be integrated as an image pre-processing in the global framework we propose.

\paragraph{Hough Transform-based detectors.}
The Hough transform~\cite{illingworth1988survey} is a traditional technique to detect lines in images.
It can be viewed as a kind of meta-template matching technique:
line evidence $L$ with $(\rho, \theta)$ polar parameters is supported by pixel observations $(x_i,y_i)$
according to the constraint that $(L): \rho = x_i \cos \theta + y_i \sin \theta$.
In practice, such technique relies on a binarization pre-processing to identify ``foreground'' pixels
which can support a set of $(\rho, \theta)$ values in polar space.
Line detection is performed by identifying maximal clusters in this space, 
which limits the accuracy the detection to a very coarse vectorization,
while relying on a high algorithmic complexity.
Several variants~\cite{mukhopadhyay_survey_2015} were proposed to overcome those limitations:
the Randomized Hough Transform~\cite{kultanen1990randomized} and Probabilistic Hough Transform~\cite{kiryati_probabilistic_1991} 
lowered the computational cost thanks to some random sampling of pixels,
and the Progressive Probabilistic Hough Transform~\cite{galamhos_progressive_1999} added the ability to predict start and end line segment coordinates, turning the approach into an effective linear object detector.
This method can tolerate gaps and overlaps to some extents,
and is highly popular thanks to its implementation in the OpenCV library~\cite{opencv_library}.
However, it is very sensitive to noise, 
to object length, %
rather inaccurate with slightly curved objects,
and cannot assign pixels to linear object instances.

\paragraph{Region growing tracers.}
The Canny edge detector~\cite{canny1986computational} is the most famous example of this category.
It is based on the following steps:
a gradient computation on a smoothed image,
the detection of local gradient maximums,
the elimination of non-maximal edge pixels in the gradient's direction,
and the filtering of edge pixels (based on gradient's magnitude) in the edge direction.
All region growing approaches improve this early algorithm, probably introduced by Burns et al.~\cite{burns_extracting_1986},
which effectively consists in finding salient edge stubs and tracing from there initial position.
LSD~\cite{von_gioi_lsd_2010} was the first approach which took linear object detection to the next level in natural images.
Fast and accurate, it is based on a sampling of gradient maximums
which are connected only if the gradient flow between these candidates is significantly strong
compared to a background noise model, according to the \emph{Helmholtz principle} proposed by \cite{desolneux_gestalt_2007}.
EDLine~\cite{akinlar_edlines_2011} and AG3line~\cite{zhang_ag3line_2021} improved the routing scheme to connect distant gradient maximums using Least Square fitting and a better sampling, accelerating and enhancing the whole process.
CannyLines~\cite{lu2015cannylines} proposed a parameter-free detection of local gradient maximums, and reintegrated the \emph{Helmholtz principle} in the routing.
Ultimately, ELSED~\cite{suarez2022elsed} further improved this pipeline to propose the fastest detector to date,
with a leading performance among learning-less methods.
These recent approaches are very fast and accurate, require no training stage, and can handle intersections.
However, they do not detect all pixels which belong to a stroke, and have limited tolerance to gaps and overlaps.
Finally, these methods require a careful tuning to be used on document images as the integrated gradient computation step tends to be problematic for thin linear objects, and leads to double detections or filters objects (confusing them with noise).

\paragraph{Deep linear object detectors.}
Region Proposal Networks, and their ability to be trained end-to-end using RoI pooling,
where introduced in the Faster R-CNN architecture~\cite{faster-rcnn}.
This 2-stage architecture was adapted to linear object detection by the L-CNN approach~\cite{zhou_end--end_2019}.
L-CNN uses a junction heatmap to generate line segment proposals, which are fed to classifier using a Line of Interest pooling,
eventually producing vector information containing start and end coordinates.
HAWP~\cite{xue_learning_2019} accelerated the proposal stage by replacing the joint detection stage with the previously-introduced Attraction Field Map~\cite{xue_learning_2019}.
F-Clip~\cite{dai2021fully} proposed a similar, faster approach using a single-stage network
which directly predicts center, length and orientation for each detected line segment.
These approaches produce very solid vector results,
but they are still not capable to assign pixels to a particular object instance.
Furthermore, their important computation and training data requirements
forces us to remove them from our current study.

\paragraph{Vertex sequence generators.}
Recently, the progress of decoders enabled the direct generation of vector data from an image input,
using a sequence generator on top of some feature extractor.
Polygon-RNN~\cite{castrejon_annotating_2017,acuna_efficient_2018} uses a CNN as feature extractor, and an RNN decoder which generates sequences of vertex point coordinates.
LETR~\cite{xu2021line} uses a full encoder/decoder transformer architecture and reaches the best wireframe parsing accuracy to date.
However, once again, such architectures are currently limited to vector predictions,
and their computation and training data requirements make them unsuitable for our current comparison.

\paragraph{Linear object trackers.}
A neglected direction, with several key advantages and which opens interesting perspectives,
was introduced in 1994 in a short series of papers~\cite{dandecy_kalman_1994,poulain_dandecy_analyse_1995,leplumey_kalman_1995}.
This approach leverages the power of Kalman filters~\cite{kalman_new_1960}, which were very successful for sensors denoising,
to stabilize the detection of linear objects over the course of two image scans (horizontal and vertical).
By tracking individual object candidates, eventually connecting or dropping them,
this approach proposed a lightweight solution, with few tunable parameters, which can \emph{segment instances} of linear objects by assigning pixels to all the objects they belong to, but also deal with noise, curved objects, gaps and noise.
However, no comparison against other approaches, nor public implementation, were disclosed.
This called for a revival and a comparison against the fastest methods to date.

\section{MOT Framework for Linear Object Detection}
\label{sec:framework}
As previously mentioned, we extend the original approach of~\cite{dandecy_kalman_1994,poulain_dandecy_analyse_1995,leplumey_kalman_1995},
which performs 2 scans over an image, plus a post-processing, to detect linear objects.
We will describe the horizontal scan only: the vertical one consists in the same process applied on the transposed image.
During the horizontal scan, the image is read column by column in a single left-to-right pass; %
each column being considered as a \emph{1-dimensional scene} containing linear object \emph{spans},
i.e. slices of dark pixels in the direction normal to linear object (\Cref*{fig:mot_line_detection}).
Those spans are tracked scene by scene, and linked together to retrieve objects with pixel accuracy.
To ensure accurate linking, and also to tolerate gaps, overlaps and intersections,
each coherent sequence of \emph{spans} is modeled as a \emph{Kalman filter} which stores information about past \emph{spans},
and can predict the attributes (position, thickness, luminance\dots) of the next most probable one.
By matching such \emph{observations} with \emph{predictions}, and then correcting the internal parameters of the \emph{filter},
observations are aggregated in a self-correcting model instance for each linear object.
Once the horizontal and vertical scans are complete, object deduplication is required to merge double detections close to 45 degrees.
Finally, using pixel-accurate information about each object instance, several outputs can be generated, and in particular:
a mapping which stores for each pixel the associated object(s),
and a simplified vector representation containing first and last span coordinates only.

\begin{figure}[tb]
  \fbox{
    \includegraphics{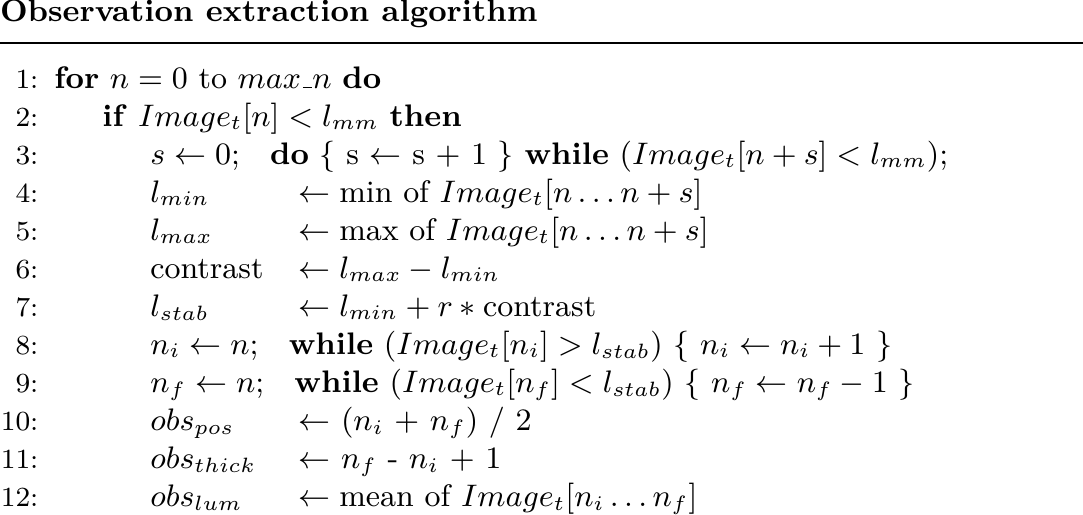}
    }

    \includegraphics[width=\linewidth]{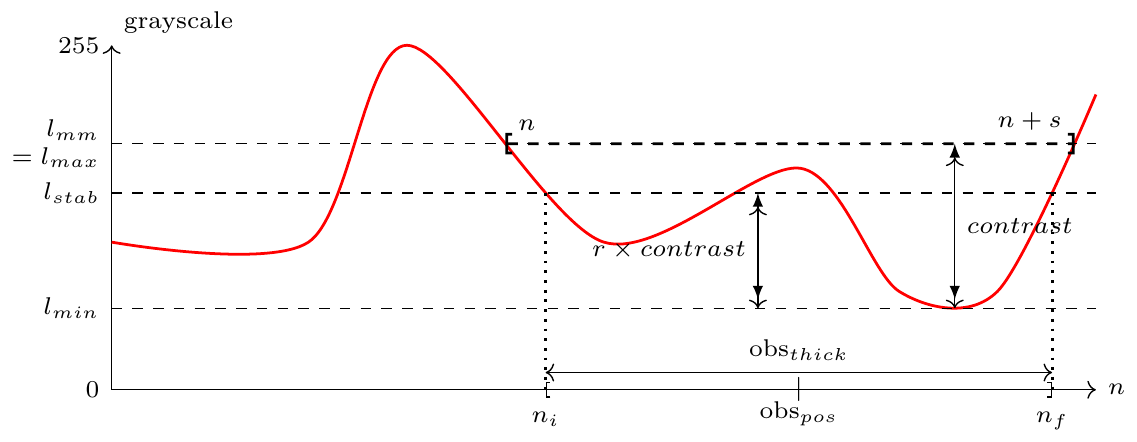}

    \caption{\small Algorithm and illustration of the observation extraction process.  The red curve represents the
      luminosity profile of a line (or a column) of pixels. The algorithm looks for a range of pixels
      $[n \ldots n+s]$ whose value do not exceed $l_{mm}$ (L.3) and computes its contrast (L.4-6). Then, this range is
      refined to extract the largest sub-range whose contrast does not exceed $r$ (a parameter of the method set to 1 in
      all our application but kept from the original method) times \emph{contrast} (L.8-9) and gives the
      \emph{observation} with features computed on lines 10-12.}
\label{fig:extraction}
\end{figure}

\paragraph{Framework overview.}
We propose to abstract this original approach into a more general Multiple Objects Tracking (MOT) framework.
This enables us to explicit each stage of such process, and introduce variants.
Linear object detection using 2-pass MOT can be detailed as follows.
\begin{description}
    \item[Pre-processing.]
    In \Cref{sec:experiments}, we will report results with grayscale images only, as this is the simplest possible input for this framework.
    However, some preprocessing may be used to enhance linear object detection.
    In the case of uneven background contrast, we obtained good results using a \textit{black top hat},
    and to be able to process very noisy images, or images with rich textures,
    it may be possible to train a semantic segmentation network which would produce some edge probability map.
    \item[Processing.]
    The horizontal and vertical scans, which can be performed in parallel, 
    aim at initializing, updating and returning a set of \emph{trackers},
    a generalization of the ``filters'' specific to Kalman's model.
    Like in the original approach, each detected instance is tracked by a unique \emph{tracker} instance.
    \emph{Trackers} are structures which possess an internal State $S$ containing a variable amount of information, according to the variant considered;
    two key methods ``predict'' and ``integrate'' which will be described hereafter;
    and an internal list of \emph{spans} which compose the linear object.
    The State and the two methods can be customized to derive alternate tracker implementations (we provide some examples later in this section). %
    For each \emph{scene}~$t$ (column or line) read during the scan, the following steps are performed. %
    Steps 1 and 2 can be performed in parallel, as well as steps 4, 5 and 6.

    \emph{1. Extract the set of observations} $O_t^j$, $j \in \left[0, \mathop{nobs}_t\right[$. %
        Observations represent linear object \emph{spans}, and contain information about their \emph{position} in the scene, \emph{luminance} and \emph{thickness}.
        At this step, some observations may be rejected because there size is over a certain threshold. This threshold is a parameter of the method.
        The algorithm and the associated illustration in \cref{fig:extraction} detail our implementation this step
        based on our interpretation of the original approach~\cite{dandecy_kalman_1994,poulain_dandecy_analyse_1995,leplumey_kalman_1995}.

    \emph{2. Predict the most probable next observation} $X_{t}^i$ for each tracker $i$,
        using its current internal State $S_{t-1}$.
        Predictions have the same structure and attributes as observations.

    \emph{3. Match extracted observations} $O_t^j$ with predicted ones $X_{t}^i$. %
        Matching is a two-step process performed for each tracker $i$.
        First, candidate observations $O_t^{i,j}$ are selected based on a distance threshold,
        and \emph{slope}, \emph{thickness} and \emph{luminance} compatibility (they must be inferior to 3 times the standard deviation of each parameter, computed over a window of past observations).
        Second, the closest observation is matched: $\hat{O}_t^i = \operatorname*{arg\,min} |O^{i,j}_t(\mathrm{position}) - X_t^i(\mathrm{position})|$.
        The same observation can be matched by multiple trackers when lines are crossing.
        Unmatched observations $\bar{O}_t$ are kept until step 5.

        \emph{4. Integrate new observations $\hat{O}_t^i$ into trackers' States $S_{t}^i$,}
        considering (in the more general case) current State $S_{t-1}^i$, scene $t$, matched observation $\hat{O}_t^i$ and prediction $X_t^i$.
        This enables each tracker to adapt its internal model to the particular object being detected.
        
        \emph{5. Initialize new trackers} from unmatched observations $\bar{O}_t$.
        New trackers are added to the active pool of trackers to consider at each scene $t$.

        \emph{6. Stop trackers of lost objects}.
        When a tracker does not match any observation for too many $t$, it is removed from the pool of active trackers.
        The exact threshold depends on the current size of the object plus an absolute thresholds for acceptable gap size.
        Those two thresholds are parameters of our method.

  \item[Post-processing.]

    \emph{Deduplication} is required because 45\textdegree-oriented segments may be detected twice. Duplications are
    removed by comparing and discarding object instances with high overlap. An optional \emph{attribute filtering} may
    then be performed, that consists in filtering objects according to their length, thickness or angle.  This
    filtering must be performed after the main processing stage to avoid missing intersection and overlaps with other
    objects, would these be linear or not: this makes possible to handle interactions between handwritten strokes and
    line segments, for instance.  Finally, \emph{outputs} can be generated by decoding the values stored in each
    \emph{tracker} object.
\end{description}

\newcommand{\myalign}[1]{%
\rlap{#1}\phantom{constrast}
}

\paragraph{Tracker variants.}
Each tracker variant features a different internal State structures (which needs to be initialized), and specific prediction and integration functions.
However, in the context of Linear Object Detection, we keep the original model of the IRISA team~\cite{dandecy_kalman_1994,poulain_dandecy_analyse_1995,leplumey_kalman_1995}:
all observations have luminance, position and thickness attributes,
and States have a similar structure, generally adding a \emph{slope} attribute to capture position change.
In this work, we consider the following variants.
To simplify the following equations, 
we refer to $t=0$ as the time of the first observation of a tracker,
and consider only one tracker at a time.%
\begin{description}
    \item[Last Observation.]
    A naive baseline approach which uses as prediction the last matched observation ($X_t = \hat{O}_{t-1}$).
    Updating its States simply consists in storing the last matched observation in place of the previous one.

    \item[Simple Moving Average (SMA).]
    Another baseline approach which stores the $k$ last matched observations and extrapolates the prediction based on them.
    This tracker uses a \emph{slope} attribute.
    Prediction for the attribute $a$ is the average of the previous observations $X_{t}(a) = (\sum_{j=t-k}^{t-1} \hat{O}_{j})/k$,
    except for the position, which is computed using the last observed position and the Exponential Moving Average (EMA) of the slope.
    Integration of new observations consists in adding them to the buffer.
    We used a buffer of size $k = \min(t,30)$ in our experiments.

    \item[Exponential Moving Average (EMA).]
    A last baseline approach very similar to the previous one, which requires only to store the last matched observation $\hat{O}_{t-1}$ and the last prediction $X_{t-1}$.
    The prediction of some attribute $a$ is computed by: $X_{t}(a) = \alpha * \hat{O}_{t-1}(a) + (1 - \alpha) * X_{t-1}(a)$ where $\alpha$ tunes importance of the new observation.
    The prediction of the position is computed using the last observed position and the EMA of the slope.
    We used a value of $\alpha = 2/(\min(t, 16) + 1)$ in our experiments.

    \item[Double Exponential~\cite{laviola2003double}.]
    This tracker uses a double exponential smoothing algorithm which is faster than Kalman filters and simpler to implement.
    $X_{t} = (2 + \frac{\alpha}{1 - \alpha}) \cdot S_{X_{t-1}} - (1+\frac{ \alpha}{1 - \alpha}) \cdot S_{X[2]_{t-1}}$ where 
    $S_{X_{t}} = \alpha \cdot O_t + (1 - \alpha) \cdot S_{X_{t-1}}$,
    $S_{X[2]t} = \alpha \cdot S_{X_{t}} + (1 - \alpha) \cdot S_{X[2]_{t-1}}$ 
    and $\alpha \in [0,1]$ (set to $0.6$ in our experiments) a smoothing parameter.

    \item[1\,\euro{} Filter~\cite{casiez20121}.]
    This tracker features a more sophisticated approach, also based on an exponential filter, which can deal with uneven signal sampling.
    It adjusts its low-pass filtering stage according to signal's derivative, and has only two parameters to configure:
    a minimum cut-off frequency that we set to $1$ in our experiments, and a $\beta$ parameter that we set to $0.007$.
    We refer the reader to the original publication for the details of this approach.

    \item[Kalman Filter~\cite{kalman_new_1960,welch_introduction_2006}, IRISA variant~\cite{dandecy_kalman_1994,poulain_dandecy_analyse_1995,leplumey_kalman_1995}.]
    This tracker is based on our implementation of the approach proposed by the IRISA team.
    The hidden State is composed of $n=4$ attributes
    $S \in \mathcal{R}^n: \begin{bmatrix}\mathrm{pos.} & \mathrm{slope} & \mathrm{tick.} & \mathrm{lum.}\end{bmatrix}\tran$,
    and the process is governed by the following equation: 
    $S_t = A S_{t-1} + w_{k-1}$
    where $A \in \mathcal{R}^{n,n}$ is the transition model
    and $w_{k-1}$ some process noise.
    Process States $S$ can be projected to observation space
    $O \in \mathcal{R}^m: \begin{bmatrix}\mathrm{pos.} & \mathrm{tick.} & \mathrm{lum.}\end{bmatrix}\tran$
    according to the measurement equation
    $O_t = H S_t + v_k$
    where $H$ is a simple projection matrix discarding the slope
    and $v_k$ the measurement noise.
    By progressively refining the estimate of the covariance matrix $Q$ (resp. $R$) of $w$ (resp. $v$),
    the Kalman filter recursively converges toward a reliable estimate of the hidden State $S$
    and its internal error covariance matrix $P$.
    \\%
    The prediction step consists in
    (1) projecting the State ahead according to a noise-free model:
    $S_{t + 1}^-  = A S_{t}$;
    and (2) projecting the error covariance ahead:
    $P_t^- = A P_{t-1} A\tran + Q$ where 
    $Q$ is the process noise covariance matrix.
    \\%
    The integration step consists in
    (1) computing the Kalman gain:
    $K_t = P_{t}^- H\tran (H P_{t}^- H\tran + R)^{-1}$
    where $R$ is the measurement noise covariance matrix;
    (2) updating estimate with measurement $\hat{O}_{t}$:
    $S_{t} = A S_{t}^- + K_t(\hat{O}_{t} - H S_{t}^-)$
    where $H$ relates the State to the measurement;
    and (3) updating the error covariance:
    $P_t = (I - K_t H) P_t^-$.
    \\%
    The following initialization choices are made, according to our experiments and the original publications~\cite{dandecy_kalman_1994,poulain_dandecy_analyse_1995,leplumey_kalman_1995}.
    We initialize each State with the values of the first observation, with a slope of 0.
    We use $P_0 = I_4$ as initial value for the error covariance matrix, $I_n$ being the identity matrix in $\mathcal{R}^n$.
    $Q$, $R$, $H$ and $A$ assumed to be constant with the following values:
    $$
        A = \begin{bmatrix} 1 & 1 & 0 & 0 \\ 0 & 1 & 0 & 0 \\ 0 & 0 & 1 & 0 \\ 0 & 0 & 0 & 1\end{bmatrix},\,
        H = \begin{bmatrix} 1 & 0 & 0 & 0 \\ 0 & 0 & 0 & 0 \\ 0 & 0 & 1 & 0 \\ 0 & 0 & 0 & 1\end{bmatrix},\,
        Q = I_4 \cdot 10^{-5},\,\mathrm{and}\;
        R = \begin{bmatrix}1 & 1 & 4\end{bmatrix}\tran.
    $$

\end{description}

\section{Experiments}
\label{sec:experiments}
We report here the results obtained by comparing the performance of \emph{training-less} approaches on two tasks:
a pure \emph{vectorization task}, where only two endpoint coordinates are required for each line segment,
and an \emph{instance segmentation task}, where pixel-accurate labeling of each line segment instance is required.
This separation is due to the limitations of some approaches which can only generate vector output,
as well as the limitations of existing datasets.
\\%
While the methods studied are \emph{training-less}, they are not \emph{parameter-free},
and such parameters require to be tuned to achieve the best performance.
In order to ensure an unbiased evaluation, we manually tuned
(because grid-search and other optimization techniques were not usable here)
each approach on the \emph{train set} of each dataset,
then evaluate their performance on the \emph{test set}.

\subsection{Vectorization Task}

\paragraph{Dataset.}
To our knowledge, no dataset for line segment detection in vector format exists for document images.
We introduce here a small new, public dataset which contains endpoints annotations for line segments
in 195 images of 19\textsuperscript{th} trade directories.
In these documents, line segment detection can be used for image deskewing.
The train (resp. test) set is composed of 5 (resp. 190) images, containing on average 4.3 line segments to detect each.
Images samples are available in the extra material.

\paragraph{Metric.}

\begin{figure}[tb]
    \centering
    \includegraphics{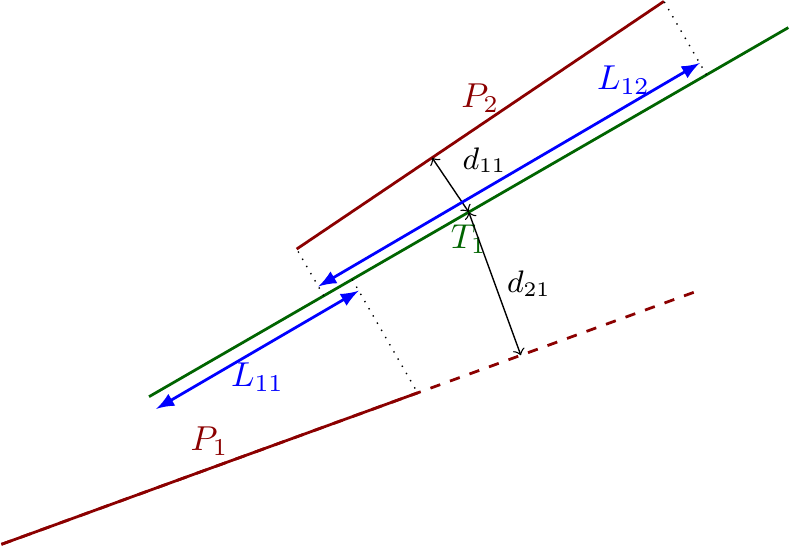}
    \caption{Matching and scoring process.
    $P_1$ and $P_2$ are associated to $T_1$ because angles are compatible
    and overlaps $L_{11}$ and $L_{22}$ are large enough.
    In this example, \emph{precision} is $\frac{|L_{11}| + |L_{21}|}{|P_{1}| + |P_{2}|}$
    and is $< 1$ because $P_1$ does not fully match the ground-truth.
    \emph{Recall} is $\frac{|L_{11} \cup L_{21}|}{|T_1|}$
    and also is $< 1$ because $T_1$ is not fully matched.%
    }
    \label{fig.eval-seg}
\end{figure}
Our evaluation protocol is a slightly modified version of the one proposed by Cho \etal~\cite{cho_novel_2018},
and is defined as follows.
Let $P = \{P_1, \cdots, P_n\}$ and 
$T = \{T_1, \cdots, T_m\}$ be respectively the
set of \emph{predicted} %
and \emph{targets} line segment (LS). %
Let $L_{ij}$ be the projection of $P_i$ over $T_j$.
We note $|X|$, the length of the LS $X$.
A predicted LS $P_i$ matches the \emph{target} $T_j$ if:
\begin{enumerate*}
    \item the prediction overlaps the target over more than 80\%  ($|L_{ij}|/|P_{i}| \ge 0.8$), 
    \item the perpendicular distance $d_{ij}$ between $T_j$' center and $P_i$ is less than 20 pixels, and
    \item the orientations of $P_i$ and $P_j$ differ at most by 5\textdegree.
\end{enumerate*}
For each $P_i$, we associate the closest LS (in terms of $d_{ij}$) among the set of \emph{matching} LSs from the targets.
We note $A = \{ A_{ij} \} \in (P \times T)$ the mapping that contains $A_{ij} = (P_i,T_j)$
whenever $P_i$ matches $T_j$.
Our protocol allows a \emph{prediction} to match a single \emph{target},
but a \emph{target} may be matched by many \emph{predictions}.
This allows for \emph{target} fragmentation (1-to-many relation).
We then compute the \emph{precision} and \emph{recall} scores as follows:
\begin{align*}
precision &= \frac{\sum_{(i,j) \in A} |L_{i,j}|}{\sum_i |P_i| } &
recall &=  \frac{|\bigcup_{(i,j) \in A } L_{i,j}|}{\sum_j |T_j|} 
\end{align*}
Roughly said, \emph{precision} stands for the quality of the \emph{predictions} to match and cover the ground-truth,
while \emph{recall} assess the coverage of the \emph{targets} that were matched,
as illustrated in \cref{fig.eval-seg}.
\\%
Our protocol differs from the original one~\cite{cho_novel_2018}
which allows \emph{prediction} to match many \emph{targets} (many-to-many relation).
This case can happen when segments are close to each other, and results in a \emph{precision} score that can exceed \emph{1.0}.
Our variant ensure precision remains in the $[0,1]$ range.
Even with this modification, these metrics still have limitations:
duplication of detections and fragmentation of targets are not penalized.
We thus propose an updated \emph{precision} as:
\begin{align*}
    precision_2 &= \frac{1}{\sum_i |P_i| } \times \sum_{(i,j) \in A} \frac{|L_{i,j}|}{|A_{(*,j)}|}
\end{align*}
with $|A_{(*,j)}| = |\{ A_{kj} \in A \}|$ being the number of matches with target $T_j$ (number of fragments).
From \emph{precision} (resp. \emph{precision\textsubscript{2}}) and \emph{recall}, the F-score (resp.
\fscoreII) is then computed and used as the final evaluation metric.

\paragraph{Results.}
We chose to only compare Hough based and Region growing algorithms featuring a rapid, public implementation and requiring no training.
\begin{table}[tb]
\setlength{\tabcolsep}{3pt}
\small
\caption{Vectorization performance and compute time of various MOT strategies on the \emph{trade directories} dataset.
F-score and \fscoreII are computed per-page and averaged on the dataset (standard deviation is shown between brackets).}
\label{table.dir-tracker}
\begin{tabular}{lrrrrr}
\toprule
{}               & {Time}       & \multicolumn{2}{c}{F-Score} & \multicolumn{2}{c}{\fscoreII}                                                     \\
{}               & {(ms)}       & {Train}                     & {Test}                    & {Train}                   & {Test}                    \\
\midrule
Last observation & \textbf{616} & \textbf{95.2 ($\pm$7.5)}    & 90.0 ($\pm$24.1)          & \textbf{92.7 ($\pm$13.0)} & 87.2 ($\pm$24.7)          \\
SMA              & 652          & \textbf{95.2 ($\pm$7.5)}    & 90.0 ($\pm$24.0)          & \textbf{92.7 ($\pm$13.0)} & 87.4 ($\pm$24.7)          \\
EMA              & 617          & 92.6 ($\pm$8.6)             & 89.7 ($\pm$24.3)          & 88.3 ($\pm$16.9)          & 86.5 ($\pm$24.8)          \\
Double exp.~\cite{laviola2003double}      & 623          & 94.6 ($\pm$7.2)             & 87.3 ($\pm$25.6)          & 85.6 ($\pm$15.9)          & 81.7 ($\pm$26.4)          \\
One euro~\cite{casiez20121}         & 627          & \textbf{95.2 ($\pm$7.5)}    & \textbf{90.1 ($\pm$24.0)} & 90.8 ($\pm$17.2)          & 87.2 ($\pm$24.7)          \\
Kalman~\cite{dandecy_kalman_1994}           & 633          & \textbf{95.2 ($\pm$7.5)}    & \textbf{90.1 ($\pm$24.0)} & \textbf{92.7 ($\pm$13.0)} & \textbf{87.6 ($\pm$24.6)} \\
\bottomrule
\end{tabular}
\end{table}

\begin{table}[tb]
\setlength{\tabcolsep}{3pt}
\caption{Vectorization performance and compute time of various line segment detection approaches on the \emph{trade directories} dataset.
Even on this rather simple dataset, these results show the superiority of MOT-based approaches for line segment detection in document images.
}
\label{table.dir-sota}
\begin{tabular}{lrrrrr}
\toprule
 {}           & {Time}       & \multicolumn{2}{c}{F-Score} & \multicolumn{2}{c}{\fscoreII}                                                     \\
 {}           & {(ms)}       & {Train}                     & {Test}                    & {Train}                   & {Test}                    \\
 \midrule
 MOT (Kalman) & 633          & \textbf{95.2 ($\pm$7.5)}    & \textbf{90.1 ($\pm$24.0)} & \textbf{92.7 ($\pm$13.0)} & \textbf{87.6 ($\pm$24.6)} \\
 \hline
 AG3Line~\cite{zhang_ag3line_2021}      & 434          & 66.2 ($\pm$23.8)            & 72.5 ($\pm$35.4)          & 25.9 ($\pm$9.9)           & 24.2 ($\pm$13.7)          \\
 CannyLines~\cite{lu2015cannylines}   & 551          & 81.2 ($\pm$22.7)            & 84.4 ($\pm$24.2)          & 39.0 ($\pm$13.5)          & 34.2 ($\pm$14.0)          \\
 EDLines~\cite{akinlar_edlines_2011}      & 314          & 83.2 ($\pm$23.6)            & 87.4 ($\pm$24.0)          & 35.5 ($\pm$8.3)           & 30.5 ($\pm$12.3)          \\
 ELSED~\cite{suarez2022elsed}        & \textbf{264} & 91.1 ($\pm$11.3)            & 87.0 ($\pm$26.6)          & 45.3 ($\pm$9.6)           & 35.2 ($\pm$13.7)          \\
 Hough~\cite{galamhos_progressive_1999}        & 419          & 80.5 ($\pm$14.5)            & 64.8 ($\pm$30.0)          & 23.5 ($\pm$9.2)           & 18.2 ($\pm$10.1)          \\
 LSD~\cite{von_gioi_lsd_2010}          & 2338         & 18.7 ($\pm$10.3)            & 12.5 ($\pm$8.5)           & 1.6 ($\pm$1.5)            & 0.5 ($\pm$0.6)            \\
 LSD II       & 2206         & 76.7 ($\pm$28.6)            & 53.3 ($\pm$43.7)          & 47.6 ($\pm$24.7)          & 20.7 ($\pm$17.9)          \\
\bottomrule
\end{tabular}
\end{table}

\begin{figure}[tb]
    \includegraphics[width=.47\linewidth]{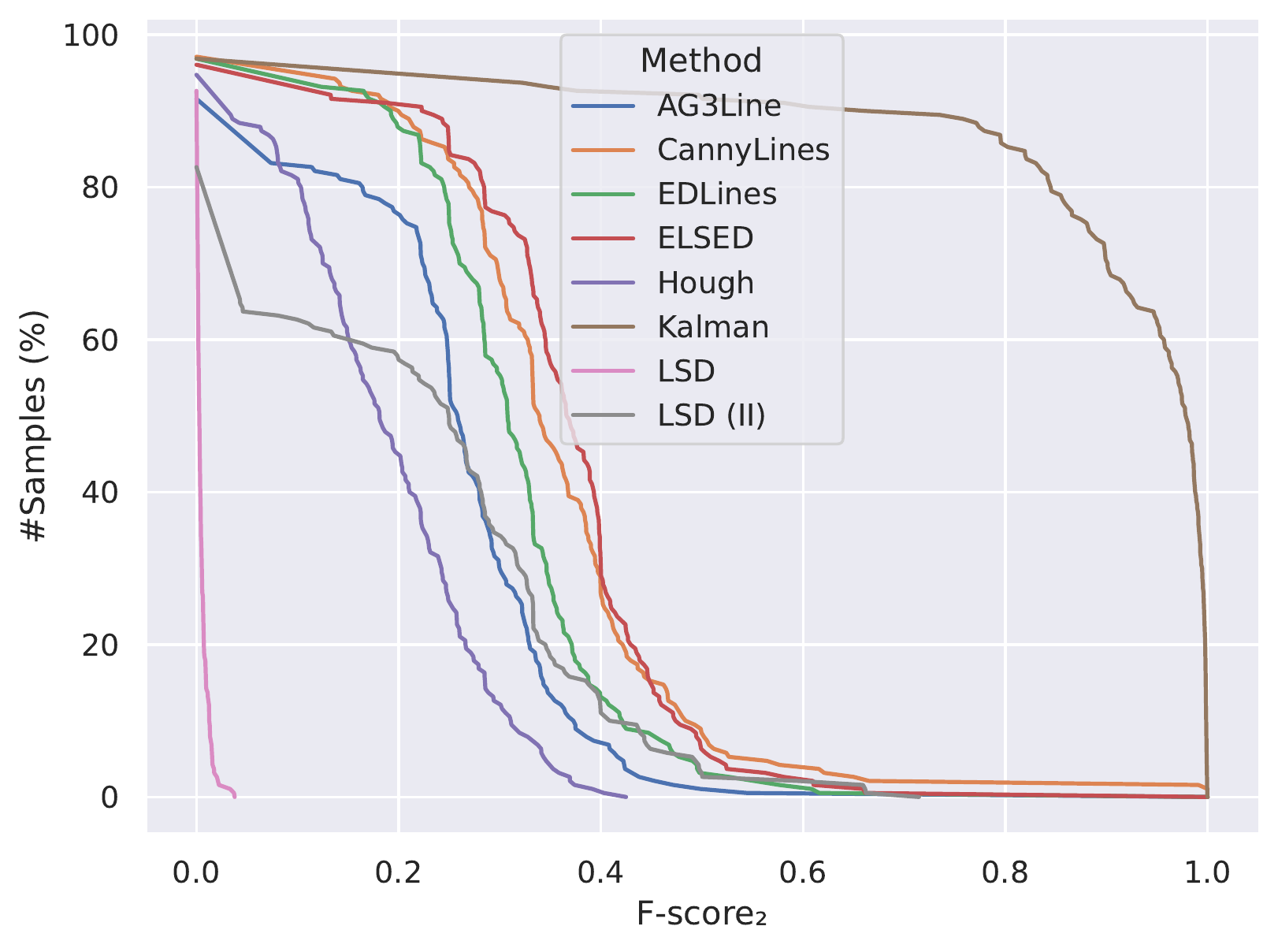}
    \hfill
    \includegraphics[width=.47\linewidth]{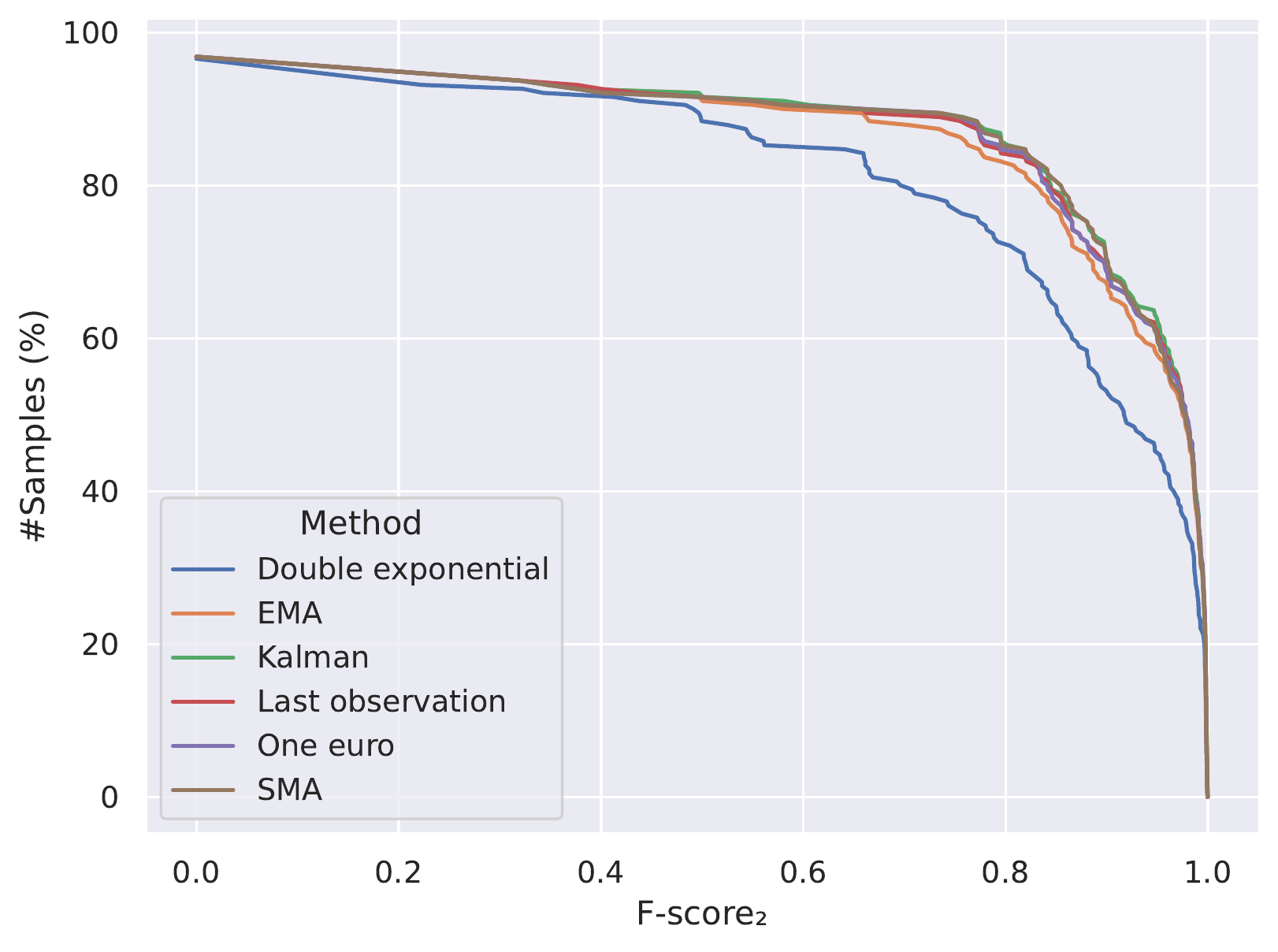}
    \caption{Distribution of the scores obtained by traditional methods (\emph{left}) and tracking systems (\emph{right}) 
    on the \emph{trade directories} dataset.
    The distribution $F(\alpha)$ counts the number of pages that gets a \fscoreII higher than $\alpha$.
    }
    \label{fig.comp-dir}
\end{figure}

In~\cref{table.dir-tracker}, we show that the original Kalman strategy performs the best on this dataset.
Nevertheless, the other tracking strategies reach almost the same level of performance
(the first five are within a 2\% range).
The differences between the F-score and \fscoreII columns are explained by
many very short detections that match a ground truth line and are more penalized with \fscoreII.
In~\cref{table.dir-sota}, state-of-the-art detectors are compared:
MOT-based (using \emph{Kalman} tracker),
Edlines~\cite{akinlar_edlines_2011},
Hough (from OpenCV)~\cite{kiryati_probabilistic_1991},
CannyLines \cite{lu2015cannylines},
LSD~\cite{von_gioi_lsd_2010},
LSD II with a filtering on segment lengths,
ELSED~\cite{suarez2022elsed},
and AG3Line \cite{zhang_ag3line_2021}.
All these techniques performs much lower than the MOT-based approach because of thick lines being detected twice
(especially with the \fscoreII metric where splits are penalized).
The two previous tables exhibit a large standard deviation of the performance
because of some bad quality pages (noise, page distortions\ldots) where most methods fail to detect lines.
This behavior is shown on~\cref{fig.comp-dir} where the \fscoreII distributions of the dataset samples are compared.
It shows some outliers at the beginning where detectors get a null score for some documents.

\subsection{Instance Segmentation Task}
    
\paragraph{Dataset.}
To demonstrate the feasibility of line segment instance segmentation, we adapted the dataset of the ICDAR 2013 music score competition for staff removal~\cite{visaniy2013icdar} to add information about staff line instances.
This was, to our knowledge, the only dataset for document images which features some form of line segmentation annotation we could leverage easily.
As the original competition dataset contain only binary information, i.e. for a given pixel whether it belong to any staff line,
we annotated the 2000 binary images from the competition test set to assign to each pixel some unique identifier according to the particular staff line it belongs to, effectively enabling to evaluate the instance segmentation performance of the MOT-based line segment detectors.
However, this enriched dataset contains only binary images with horizontal, non-intersecting staff lines.
To enable hyperparameter calibration, we randomly selected 5 images to use as a train set, and kept the 1995 others in the test set.

\paragraph{Metrics.}
We use the COCO Panoptic Quality (PQ) metric~\cite{kirillov_panoptic_2019} to measure instance segmentation quality.
The metric is based pairings between proposed and ground-truth regions which overlap over more than 50\%.
The Intersection over Union (IoU) is used to score the pairings and enables to compute an instance segmentation quality (``COCO PQ'') that we report here.
Even if COCO PQ has become the standard metric for instance segmentation evaluation, it has two main limits:
1. it does not measure fragmentation as it considers only 1-to-1 matches;
2. IoU is somehow unstable on thin and long objects.
For the sake of completeness, we also report results from the binary segmentation metric used in the original ICDAR 2013 competition~\cite{visaniy2013icdar}.
The problem is seen as binary classification of pixels from which an F-Score is computed.
This evaluation is not really relevant in our case because it does not include any instance information,
but is shows that MOT-based techniques can be used for binary detection.

\paragraph{Results.}
\begin{figure}[tb]
\includegraphics[width=.45\linewidth]{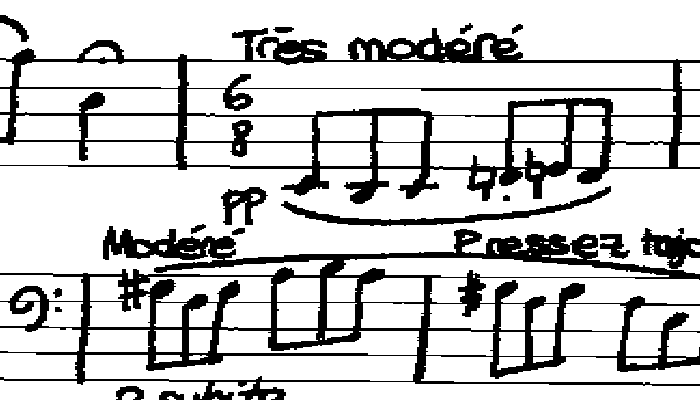}
\hfill
\includegraphics[width=.45\linewidth]{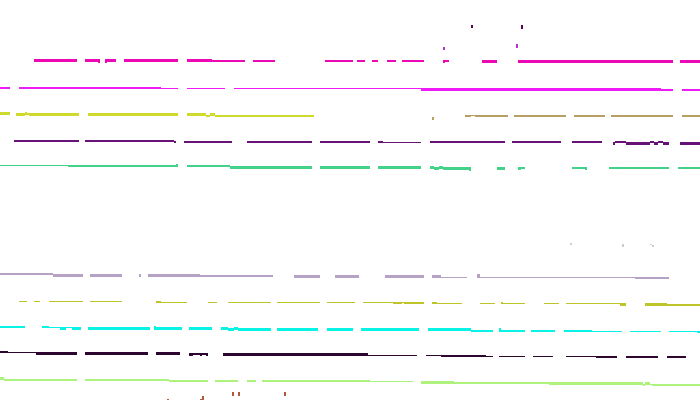}
\caption{Excerpt of the original image of staff lines (700 x 400) and its instance segmentation performed by the Kalman Filter
predictor in 60ms.%
}
\label{fig.seg.stafflines.kalman}
\end{figure}
\begin{table}[tb]
\setlength{\tabcolsep}{3pt}
\caption{Instance segmentation performance and compute time of MOT-based approaches on the \emph{music sheets} dataset adapted from ICDAR 2013 music score competition~\cite{visaniy2013icdar}.
For completeness, we also report the winner of the binary segmentation contest, though such method only performs a semantic segmentation.
}
\label{table:ms-tracker}
\begin{tabular}{lrrrrr}
\toprule
{}           & {Time}      & \multicolumn{2}{c}{Panoptic Quality} & \multicolumn{2}{c}{F-Score (ICDAR'13)}                                            \\
{}                 & {(ms)}              & {Train}                              & {Test}                    & {Train}                   & {Test}                    \\
\midrule
Last observation   & 323             & 86.3 ($\pm$ 5.2)                     & 83.7 ($\pm$ 11.1)         & 95.9 ($\pm$ 2.1)          & 95.4 ($\pm$ 2.7) \\
SMA                & 323             & 67.6 ($\pm$ 17.0)                    & 66.0 ($\pm$ 17.6)         & 90.7 ($\pm$ 6.5)          & 89.9 ($\pm$ 7.4)          \\
EMA                & 322             & 74.0 ($\pm$ 14.9)                    & 65.5 ($\pm$ 18.4)         & 92.5 ($\pm$ 4.7)          & 89.6 ($\pm$ 7.7)          \\
Double exp.~\cite{laviola2003double}        & \textbf{320}    & 55.4 ($\pm$ 16.2)                    & 51.7 ($\pm$ 15.8)         & 87.3 ($\pm$ 5.0)          & 83.8 ($\pm$ 8.6)          \\
One euro~\cite{casiez20121}           & 327             & \textbf{87.2 ($\pm$ 5.9)}            & \textbf{85.1 ($\pm$ 9.6)} & \textbf{95.9 ($\pm$ 2.1)} & \textbf{95.7 ($\pm$ 2.2)} \\
Kalman~\cite{dandecy_kalman_1994}             & 328             & 85.0 ($\pm$ 7.1)                     & 80.7 ($\pm$ 15.6)         & 95.3 ($\pm$ 2.5)          & 94.1 ($\pm$ 5.6)          \\
\hline
LRDE-bin~\cite{visaniy2013icdar}                &              &                    &          &           & \textbf{97.1}          \\
\bottomrule
\end{tabular}
\end{table}

\Cref{table:ms-tracker} shows the performance of the various trackers to identify staff lines.
All trackers successfully retrieve the staff lines.
The score difference is explained by the way the trackers handle line intersections
(when a staff line is hidden behind an object).
The \emph{one-euro} and \emph{last observation} tracking strategies
perform almost equally and are the ones that handle the best such cases.

\begin{figure}[tb]
    \centering
    \fbox{\includegraphics[width=.30\linewidth]{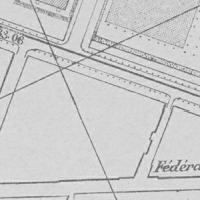}}
    \fbox{\includegraphics[width=.30\linewidth]{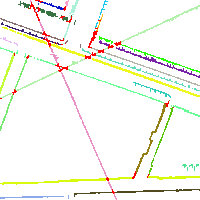}}
    \fbox{\includegraphics[width=.30\linewidth]{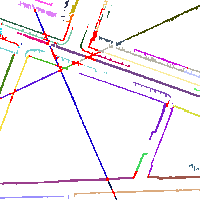}}
    \caption{Instance segmentation on map images.
    Left: original image, center: Kalman tracker, right: EMA tracker.
    With the same filtering parameters, the EMA tracker is more sensitive to gaps and overlaps and fragments objects.}
    \label{fig.segmap}
\end{figure}

\paragraph{Qualitative evaluation on the municipal atlases of Paris.}
Finally, to assess the capabilities of MOT-based line segment detectors,
we processed a selection of images from the dataset of the ICDAR 2021 competition on historical map segmentation~\cite{chazalon.21.icdar.mapseg}.
The challenges of this competition relied on an accurate detection of 
boundaries of building blocks, of map content on the sheet, and of georeferencing lines;
all of them being mostly linear objects in the map images.
We retained 15 images of average size 5454x3878 (21~Mpix) for our experiment:
3 for the training set, and 12 for the test set.
Because no pixel-level ground truth exists for these images, we report qualitative results.
The computation times of the trackers are similar: about 5-6s per 21 Mpix maps image.
Time variations are due to the number of trackers to update during the process
that depends on the observations integrated with the tracking strategy.
From a quality standpoint (outputs available in extra material),
the Kalman strategy outperforms all the other tracking strategies as shown in~\cref{fig.segmap}.
Indeed, these documents contain many overlaps and noise that create discontinuities when extracting observations.
Kalman filters enable recovering the lines even if there are hidden behind some object.
On the other hand, the other trackers with simpler prediction models
(such as the \emph{one-euro} or the \emph{last observation} trackers)
integrate wrong observations and lead to line segment fragmentation.

\section{Conclusion}
Our goal was to implement a line segment detection method that was first proposed in the 1990s~\cite{dandecy_kalman_1994,poulain_dandecy_analyse_1995,leplumey_kalman_1995}.
However, we generalized it within the Multiple Object Tracking framework that we proposed.
We demonstrated the efficiency of the original proposal, which was based on Kalman filters, and also suggested some competitive alternatives.
These approaches are highly robust to noise, overlapping contents, and gaps. They can produce an accurate instance segmentation of linear objects in document images at the pixel level.
Results can be reproduced using open code and data available at \url{https://doi.org/10.5281/zenodo.7871318}.

\subsubsection{Acknowledgements.}
This work is supported by the French National Research Agency under Grant ANR-18-CE38-0013 (SoDUCo project).
\clearpage
\bibliographystyle{splncs04}
\bibliography{bibliography}
\end{document}